\documentclass{article}
\usepackage{spconf,amsmath,graphicx,textcomp}
\usepackage{xcolor}
\usepackage{url}
\usepackage{hyperref}

\title{STREAMING, FAST AND ACCURATE ON-DEVICE INVERSE TEXT NORMALIZATION FOR AUTOMATIC SPEECH RECOGNITION}
\name{\begin{tabular}{c}Yashesh Gaur, Nick Kibre, Jian Xue, Kangyuan Shu, Yuhui Wang, Issac Alphanso, Jinyu Li, Yifan Gong\end{tabular}}
\address{Microsoft Corp., U.S.A}
\copyrightnotice{978-1-6654-7189-3/22/\$31.00~\copyright2023 IEEE}
\begin{document}
%\ninept
%
\maketitle
\begin{abstract}

Automatic Speech Recognition (ASR) systems typically yield output in lexical form. However, humans prefer a written form output. To bridge this gap, ASR systems usually employ Inverse Text Normalization (ITN). 

In previous works, Weighted Finite State Transducers (WFST) have been employed to do ITN. WFSTs are nicely suited to this task but their size and run-time costs can make deployment on embedded applications challenging.

In this paper, we describe the development of an on-device ITN system that is streaming, lightweight \& accurate. At the core of our system is a streaming transformer tagger, that tags lexical tokens from ASR. The tag informs which ITN category might be applied, if at all. Following that, we apply an ITN-category-specific WFST, only on the tagged text, to reliably perform the ITN conversion. We show that the proposed ITN solution performs equivalent to strong baselines, while being significantly smaller in size and retaining customization capabilities.

\end{abstract}
\begin{keywords}
Inverse Text Normalization, Automatic Speech Recognition, on-device, streaming.
\end{keywords}
\section{INTRODUCTION}
\label{sec:intro}

Inverse Text Normalization (ITN) \cite{Sproat2001NormalizationON} is an important component of many speech recognition applications. It converts the ASR output, which is usually in a lexical (or verbalized) format, to display (or written) format. This makes the ASR transcription much more suitable for human consumption and downstream Natural Language Understanding (NLU) tasks. For instance, the ASR output might say \textit{``that will be four fifty"}, but it would be preferable for it to read \textit{``that will be 450"}. However, note that the lexical forms are generally more ambiguous than their display-formats. In the above example, depending on the context, \textit{``four fifty"} could have been time (``4:50"), currency (``\$4.50") or simply a number (``450"). It is the job of the ITN model to study the context and resolve conflicts. The task of ITN is inherently ambiguous. In this work, we aim to solve for a wide variety of ITN categories, including but not limited to \texttt{time}, \texttt{money}, \texttt{url}, \texttt{email}, \texttt{phone}, \texttt{math}, \texttt{address}, \texttt{abbreviation}, \texttt{ordinal} etc. In total, we cater to 16 different ITN categories. 

There have been relatively small number of works that address the problem of ITN. In the context of a ``hybrid" ASR, \cite{Sak2013WrittendomainLM} built a language model directly in the written domain. To that end, they built a letter-to-sound model that could go from display-format to lexical-format. A similar display-format language model was built by \cite{Ju2008ALA} for voice search scenarios. There have also been works published on Text Normalization (TN), which is the opposite of ITN. Since the rules for TN can be inverted and leveraged for the process of ITN, this task is also of relevance. \cite{schlippe10_interspeech} formulates TN as a statistical machine translation problem and \cite{Sproat2016RNNAT} advances the same approach using recurrent neural networks and attention mechanisms. Along the same direction, more recently, \cite{sunkara2021} used transformer-based seq2seq models to perform ITN. They also used FSTs to recover from catastrophic errors in beam search. Another type of modeling solution for ITN was proposed in \cite{shugrina-2010-formatting} where authors relied on grammar rules to generate multiple display-format hypothesis, which were re-ranked using a class-based language model. Building on this work, \cite{issac_slt} also proposed display-format candidate generation using rules encoded within a WFST and re-ranking  them using Gradient Boosted Decision Tree ensemble (GBDT). In our experience, the framework of candidate generation and context-based re-ranking is very effective. As described in Section \ref{sec:wfst} and \ref{sec:model_details_wfst}, we use this framework as one of our baselines.

In this work, we adopt a different approach to ITN. We break the problem of ITN down into 2 disjoint steps: Tagging \& Transduction. More concretely, our approach consists of a new transformer-based tagger, which tags incoming lexical tokens from ASR in a streaming manner. The tags contain information about the ITN category that can be associated with any tagged span. Once, a tagged span is available, an ITN-category-specific WFST performs the actual conversion, only on the tagged part of a sentence. The proposed approach has many advantages that are discussed in Section \ref{proposed-approach} and demonstrated in Section \ref{sec:results}. The contributions in this paper are as follows: \textbf{\textit{(i)}} We propose a novel modeling solution for ITN. It splits the task into tagging \& transduction. This allows us to get high quality, streaming and light-weight models that can be deployed to on-device applications. \textbf{\textit{(ii)}} We design a chunk-based transformer tagger, which enables streaming ITN. We show that this design can be configured to trade-off between accuracy and latency. \textbf{\textit{(iii)}} Our design proposes a tag to denote a particular ITN category. For each ITN category, we have an FST that encodes the corresponding the rules of conversion. The FST can be easily swapped with an updated one. This enables much easier, on-the-fly customization, without any retraining. \textbf{\textit{(iv)}} We present an extensive empirical study on large-scale data. We benchmark the proposed ITN solution and show that it's quality is on par with strong baselines, that are an order of magnitude larger in size. We also demonstrate significant advantages in terms of computational costs.

% \begin{enumerate}
%     \item We propose a novel modeling solution for ITN. It splits the task into tagging \& transduction. This allows us to get high quality, streaming and light-weight models that can be deployed to on-device applications.
%     \item We design a chunk-based transformer tagger, which enables streaming ITN. We show that this design can be configured to trade-off between accuracy and latency.
%     \item Our design proposes a tag to denote a particular ITN category. For each ITN category, we have an FST that encodes the corresponding the rules of conversion. The FST can be easily swapped with an updated one. This enables much easier, on-the-fly customization, without any retraining. 
%     \item We present an extensive empirical study on large-scale data. We benchmark the proposed ITN solution and show that it's quality is on par with strong baselines, that are an order of magnitude larger in size. We also demonstrate significant advantages in terms of computational costs.
% \end{enumerate}

To the best of our knowledge, the closest related work to our approach was in \cite{Pusateri2017AMD}. In \cite{Pusateri2017AMD}, the authors train a bidirectional LSTM that labels lexical text. These labels specify a sequence of edits and transformations to perform on the lexical-format text. Our work differs from this work in many aspects. Firstly, our chunk-based transformer design enables our solution to work in streaming manner, with configurable latency. In contrast, the solution in \cite{Pusateri2017AMD} is strictly offline. Secondly, our tags reflect ITN categories and not specific edits. Since rule updates happen on the level of ITN-categories, it makes customization very convenient, without retraining any neural components. Lastly, on large scale-data, we show that our method can outperform strong baselines that are many times larger in size. In the next section, we discuss several modeling solutions that were considered for on-device ITN.

\section{ON-DEVICE INVERSE TEXT NORMALIZATION}
\label{sec:itn}

\subsection{Modeling ITN within E2E ASR}\label{speech2itn}

Conventionally, ITN has been a modeled as separate component that resides in the post-processing pipeline of the ASR. However, the end-to-end (E2E) training paradigm \cite{li20_interspeech, 8268937} allows us to train a model that goes from speech to display-format text directly. This means that ITN is learnt implicitly within the ASR model. This has several benefits. Firstly, ITN tokens come out in a truly streaming fashion with no additional latency. Moreover, the memory footprint of the model is reduced since there is no external ITN model to store. Beam search during inference also becomes less complex if we do not consider an external ITN system.

However, learning ITN within an E2E-trained ASR model also has some drawbacks. Since speech recognition is tightly coupled with ITN in this scenario, we lose all the flexibility in ITN system configuration. This is important because written form varies by domain and geography. In the absence of external and configurable ITN, one would need to train a different E2E-ASR model for every domain. Updating any ITN rules would also need retraining of the ASR model on all the data, which is a very time and energy consuming task. Moreover, because speech to written-format mapping is more complex than speech to verbalized text. Hence, an E2E speech-to-display-format ASR can result in reduction of overall model accuracy. Unlike tasks like punctuation, ITN does not have any evidence in the acoustic signal. This suggests that modeling ITN externally, makes more sense, even for E2E ASR. Such a model will take lexical output of ASR as input and convert that to display form. The following sections describe the modeling solutions we tried to model ITN separately.

% \subsection{Modeling ITN separately (2-stage approaches)}\label{itn-2-Stage}

\subsection{Weighted Finite State Transducers with rescoring}\label{sec:wfst}

WFST \cite{Mohri2004} are very nicely suited to the task of ITN because they allow application of arbitrary hand-crafted rules and, in many scenarios, can perform the task in a compact and robust manner. Since spoken-forms are ambiguous, we use the spoken-to-written FSTs to map them to multiple tagged written-form candidates, and use a ranker to choose the optimal one depending on the context. The ranker is utilized in a simple log-linear interpolation setup. We choose an n-gram model as our main choice for a ranker. This n-gram is also built into the FST. We also consider employing an additional LSTM ranker to further improve contextual re-ranking. The conversion rules in the FST are created \&  maintained by a pool of experts. This design is inspired by previous works like \cite{shugrina-2010-formatting, issac_slt}. %and it has proven to work well in our internal evaluations.

\subsection{Modeling ITN as a Seq2Seq task}\label{sec:itn-s2s}

In this work, we also consider the transformer-seq2seq architecture \cite{vaswani2017attention}, to model the task of ITN. Learning the ITN task in an end-to-end manner has many advantages. Firstly, unlike WFSTs, where experts need to prepare the conversion rules, these models learn all the rules entirely from data, with no involvement from human experts. Hence, it is much more scalable to new domains and languages, as long as sufficient amount of data is available. Secondly, since these models tend to be all-neural, their size can be compressed using a myriad of techniques \cite{li2014learning, Wang2022OnlineKD,Srinivas2017TrainingSN,Zhang2018StabilizingGF,Fan2020ReducingTD,Molchanov2016PruningCN,Han2016DeepCC}. The WFST models, on the other hand, tend to blow up in size when trained on large amount of data and can't be compressed effectively.

Despite these advantages, there are a few big challenges that prohibit the deployment of an all-neural model for ITN. The first big challenge is customization. ITN models are required to change their behavior by ingesting arbitrary human-specified rules, say from a configuration file. This kind of functionality is difficult to enable with all-neural models since they require a large amount of data to learn. The second major challenge is potential lack of robustness. Even when trained with large amounts of data, an all-neural ITN model can still suffer from poor generalization. A good example is that of phone numbers. There is a combinatorial complexity associated with the total number of 10-digit numbers that can form a phone number. Even when large training data is used, there is still a good chance that we will get a phone number that happens to be out-of-distribution, and is consequently formatted incorrectly. One can try an augment the training data to mitigate this issue, but this is not a scalable solution since such combinatorial explosion also happens for other categories like postal codes, money, fractions, time etc. Note that the above challenges are where the WFSTs show their strength. As long as rules can be expressed as regex, WFST can apply any conversion in a very robust manner.

\subsection{Proposed approach: Transformer Tagger + WFST}\label{proposed-approach}

\begin{figure}[t]
  \centering
  \includegraphics[width=\linewidth]{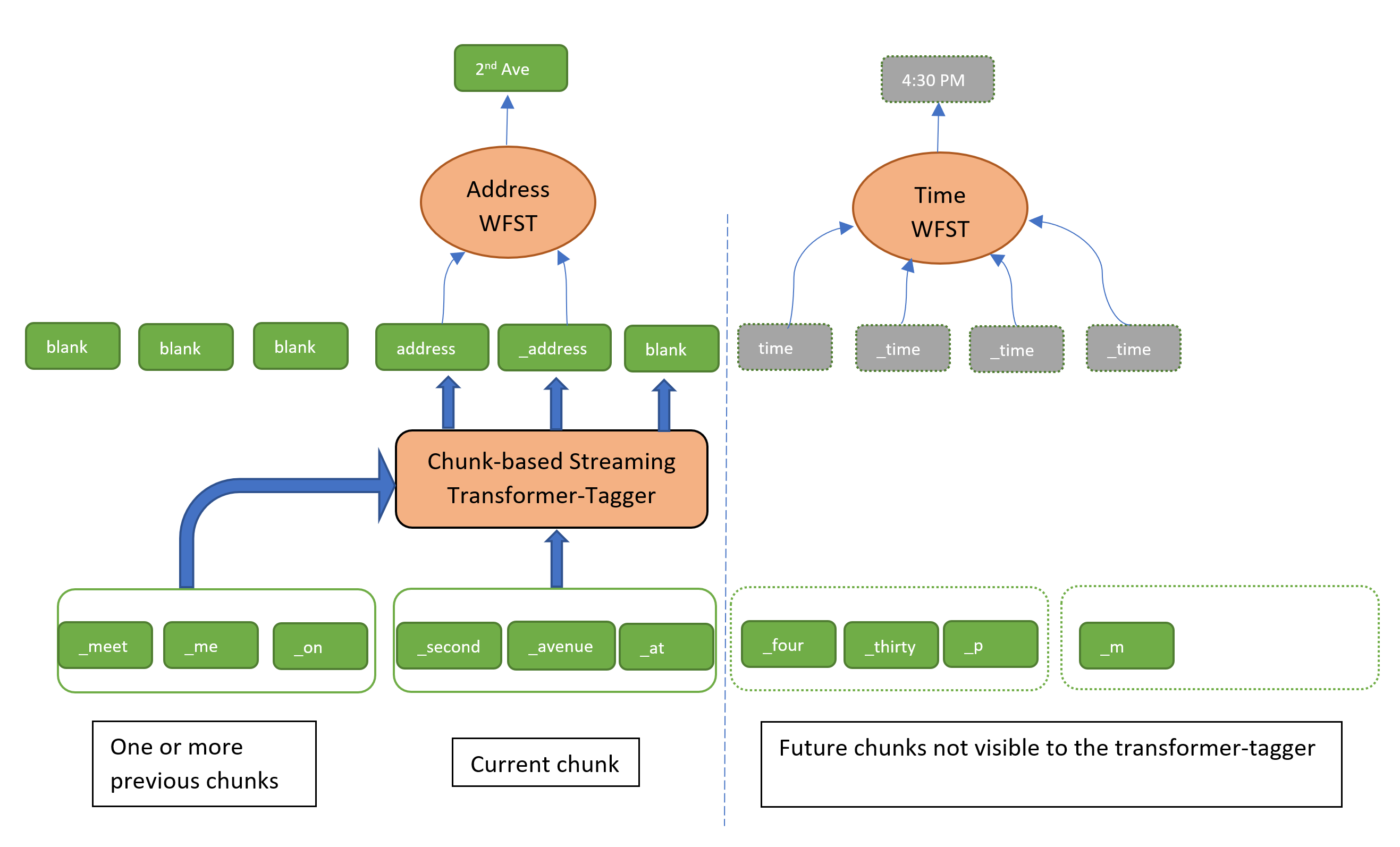}
%   \vspace{-7mm}
  \caption{High-level architecture of the proposed approach.}
  \label{fig:on_device_itn}
%   \vspace{-5mm}
\end{figure}

Our proposed approach is built on the insight that the task of ITN can be broken down into 2 disjoint steps. The first step is to figure out which parts of the sentence needs to be converted and the corresponding ITN category for them. The second step is the actual conversion according to the rules of the ITN category. Hence, we propose a “hybrid” method to ITN, which combines the robustness and customizability of a WFST and the strong predictive powers of a neural network like Transformer. More specifically, we train a Transformer neural network “tagger” that processes the output of ASR in a streaming manner and predicts a “tag” for every input token. Each tag is associated with a certain ITN category. For tokens that do not belong to any ITN category, a “blank” token is put out. Once the tags are predicted, the WFST is responsible for the actual conversion. The WFST component is a collection of several FSTs, where each FST is responsible for a particular ITN category or tag. This also helps reduce the overall size of the FST component. Please note that a particular FST is applied only to the corresponding tagged span, and not the full utterance. The approach is depicted in figure \ref{fig:on_device_itn}.

% \subsubsection{A streaming Transformer Tagger}\label{transformer_tagger}

% The tagger model's responsibility is to predict \textit{(i)} if a token needs to be translated from lexical format to display format \textit{(ii)} if so, which category-specific rules needs to be used for conversion. 
The transformer-tagger learns to use the context, both history and limited future, to predict what tag needs to be assigned to any input token. To make the transformer architecture work in a streaming manner, we use a “chunk-based” processing scheme. Similar schemes have also been utilized to make Transformer blocks within ASR encoders, work in a streaming manner \cite{9413535, zhang2020transformer}. More specifically, the transformer only processes a certain chunk of tokens at a time, and it does not have access to all the tokens in the future to make the prediction. It can however consider the chunks in the past. This chunk-based processing setup is illustrated in figure \ref{fig:on_device_itn}.

% \begin{figure}[t]
%   \centering
%   \includegraphics[width=\linewidth]{transformer_tagger.png}
% %   \vspace{-7mm}
%   \caption{Architecture of a transformer-tagger working in a streaming manner.}
%   \label{fig:transformer_tagger}
% %   \vspace{-5mm}
% \end{figure}

% \subsubsection{Chunk size, latency \& ITN performance}\label{sec:chunk_size_vs_latency_vs_accuracy}

Chunk-based processing means that output is not available until all the tokens in a chunk are available. This means that some algorithmic latency is incurred at every token. Consider the example in Figure \ref{fig:on_device_itn}. The chunk size is 3. It's clear that 1\textsuperscript{st}, 2\textsuperscript{nd} and 3\textsuperscript{rd} token incur a latency of 2, 1 and 0 tokens respectively, bringing the average latency to 1 token. It follows that a larger chunk size means a larger algorithmic latency. For streaming and on-device applications, a very small latency is preferred. This would imply that a smaller chunk size is preferred. However, a smaller chunk size means limited look ahead and a smaller window to do context modeling. Hence, there is a trade-off between latency and accuracy.

% \subsubsection{Tagged Span Mapping via WFSTs}\label{sec:itnfsts}

\begin{figure}[t]
  \centering
  \includegraphics[width=\linewidth]{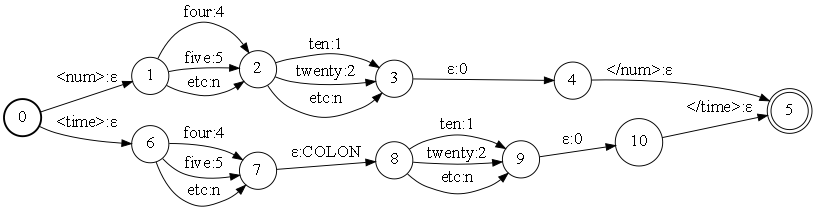}
%   \vspace{-7mm}
  \caption{Example of a Tag-Consuming ITN FST Network}
  \label{fig:itnfst}
%   \vspace{-5mm}
\end{figure}

After the tagger assigns labels to lexical recognition results, tagged spans of words are mapped to symbols via a FST.  This FST consumes both lexical words and XML-style tags to indicate categories. For instance, if ``at four thirty" is tagged \texttt{[blank, time, \_time]}, then \texttt{\textless{}time\textgreater{} four thirty \textless{}/time\textgreater{}} will be passed to the FST and mapped to ``4:30" (see Fig. \ref{fig:itnfst} for an example). This FST is compiled from a pattern-matching \& rewriting rule language, which supports transformations (e.g., ``5:five"), insertions (``$\epsilon$:word") \& deletions (``word:$\epsilon$"). We combine these transformations via regex-like notation, and references to subrules.  %Here is a simplified sample (rules are written in the display-to-lexical direction, and inverted for building ITN FST's):

% \begin{verbatim}    
% def posdig (1:one|2:two|3:three|...)
% def teen (0:ten|1:eleven|2:twelve|...)
% def ty (2:twenty|3:thirty|4:forty|...)
% def num ( $posdig | 1:<eps/> $teen \
%  | $ty (0:<eps/>|$posdig))
% def money \$:<eps/> $num <eps/>:dollars
% top <eps/>:<num> $num <eps/>:</num> \ 
%  | <eps/>:<money> $money <eps/>:</money>
% \end{verbatim}

\section{EXPERIMENTAL DETAILS} \label{sec:pagestyle}

% In this section, we explain the model details and data setup for the various modeling solutions we tried. We also explain the evaluation setup \& metrics.

\subsection{Model Details and Training Data} \label{sec:model_details}

\subsubsection{Lexical \& Display format ASR} \label{sec:model_details_asr}

The model architecture for our on-device ASR is a Transformer-Transducer (T-T) \cite{9413535}. The encoder in our T-T model works in a chunk-based manner which allows streaming operation. It contains 24 transformer blocks. Each block contains 16 heads, 4096 dimensional FFN (Feed Forward Network) and a convolutional front-end with 512 channels. The decoder is stack of 2-layer unidirectional LSTMs. Number of hidden units in each LSTM is 1024, which is projected down to 512. The Joint network is a fully connected MLP with dimension 1024. We train all the components in an end-to-end manner using the transducer loss \cite{graves2012sequence}. The optimizer of choice is adamW \cite{Loshchilov2019DecoupledWD} which follows the Noam \cite{vaswani2017attention} learning rate (LR) schedule. For more details on model architecture and training, please refer to \cite{9413535}.

The ASR model is trained on 50k hours of internal speech data. This data is anonymized with all Personally Identifiable Information (PII) removed. The speech data is featurized to 80 dimensional Mel Frequency Cepstral Coefficients (MFCC) and the target sentence is tokenized using using a vocab of 4k word pieces \cite{kudo-richardson-2018-sentencepiece}. The target transcription are originally present in lexical format. Training on this data gives us the ``lexical-ASR". To train ``display-format-ASR", we convert our lexical-format transcripts to display-format. To accomplish that, we leverage another powerful ITN solution that is not constrained in memory or compute like the on-device ITN model. We do not explain it's model architecture or training setup due to limited space in the paper, but have benchmarked it to be of high quality. After the conversion, we form a new 4k vocab \cite{kudo-richardson-2018-sentencepiece} from the display-format transcriptions and train the ASR model like before. Both lexical \& display-format ASR have the same number of parameters ($\sim$80M).

\subsubsection{Weighted Finite State Transducers with rescoring} \label{sec:model_details_wfst}

% As described in Section \ref{sec:wfst}, WFST with rescoring is an important baseline for us since it has been shown to work really well. This system overgenerates ITN candidates and leverages language models to rerank the converted outputs based on the context.

A major part of the FST system is based on hand-written rules, for \textit{e.g.} \texttt{numberword-to-digit} mappings.  The FST system also embeds a first scoring stage, an n-gram (N=7) LM. This n-gram is trained on a relatively small set ($\sim$ 2M strings) carefully curated to help choose between competing ITN mappings.  For example, since there are sequences of words that are often ambiguous between times and numbers (\textit{e.g., ``four thirty"}), we make sure that the training corpus for the n-gram LM has many examples of \texttt{Time}, in a variety of contexts, and also many examples of strings whose lexical forms could be a \texttt{Time}, but contextually should not be (\textit{i.e.}, both \textit{``see you at 4:30"} and \textit{``on route 430"}).

The FST system can generate nbest alternates. This allows to use an additional LSTM model to rerank better. We can simply use a  weighted combination of the FST score and the LSTM score. In our experiments, the LSTM model is trained on a set of $\sim$ 700M strings. The data is drawn from a coprora of email text, a collection of email addresses, common crawl data, and search queries. The selection of data from common crawl is loosely constrained to ITN-relevant strings, with some heuristics to avoid ``mis-formatted" data such as \textit{``see you at 430"}.  %Search queries were selected to target specific domains, especially addresses, since these are often ambiguous and since postal codes can be a particular challenge to ITN.

\subsubsection{Seq2Seq models} \label{sec:model_details_s2s}

Following Section \ref{sec:itn-s2s}, we train a transformer-based seq2seq model \cite{vaswani2017attention} that learns to translate lexical-format into display-format. The encoder of this model has 6 transformer blocks. Each block has 8 attention heads, 256 dimensional FFN \& 256-dimensional input embedding. The decoder is a 2-layer transformer, with the same number of attention heads, FFN embedding size and embedding dimension as the encoder. The total number of parameters in this model is $\sim$6M. We train this model in an end-to-end way using a cross-entropy loss. To train, we used the adamW optimizer \cite{Loshchilov2019DecoupledWD} with `noam'  learning rate decay scheme \cite{vaswani2017attention}. At the end of the training, we pick the checkpoint with the best performance on a held out set. We also train a proportionally scaled-up version of the above model. It has $\sim$82.5M parameters. We refer to 1\textsuperscript{st} and 2\textsuperscript{nd} model as `S2S-small' \& `S2S-large' respectively.

We train both S2S models on a dataset that has $\sim$240M parallel lexical \& display-format sentences. The data was sourced from our company's speech service transcriptions, first-party partners and free online data sources. We mined internal structured databases in order to extract address queries, popular URLs (aggregrated over six months using page view counts) and common search queries (aggregrated over six months using impresson counts). The common crawl data was filtered to extract specific patterns like plurals, ordinals, date, time and currency. Finally, we used anonymized text messages from our company's phone keystroke application to augment the training set. All the data was cleaned and PII was removed. 

% \begin{table}[t]
%   \caption{The data used to train the inverse text normalization sequence-to-sequence transformer models.}
%   \label{tab:t2t_results}
%  \vspace{3mm}
%   \centering
%   {\footnotesize
%   \begin{tabular}{lc|ccc}
%     \hline
% Set &  Source & Utterances \\ \hline
% Queries & Bing & 6,178,642 \\
% URLs & Bing & 47,538,054 \\
% Addresses & Bing & 9,281,537 \\
% Currency & Common Crawl & 33,684,482 \\
% Time & Common Crawl & 355,680 \\
% Dates & Common Crawl & 5,000,000 \\
% Number Ranges & Common Crawl & 1,187,857 \\
% Plural Numbers & Common Crawl & 2,343,161 \\
% Ordinals & Common Crawl & 3,203,542 \\
% Email & Common Crawl & 121,891 \\
% Transcriptions & Speech Services & 1,153,069 \\
% Messages & WPKS & 126,916,923 \\ \hline

%   \end{tabular}
%   }
%   \vspace{-3mm}
% \end{table}

\subsubsection{Transformer Tagger + WFST (Proposed approach)} \label{sec:model_details_tagger_wfst}

The transformer tagger follows a similar architecture to that of the ASR encoder. It has 6 transformer blocks, with each having a 256-dimensional FFN, 8 heads and 256-dimensional input embedding. It works in a chunk-based manner to make the tagging process streaming. For each token, the tagger learns to predict one out of 19 tags using a cross entropy loss. We use a dropout of 0.1 throughout the network

\begin{figure}[t]
  \centering
  \includegraphics[width=\linewidth]{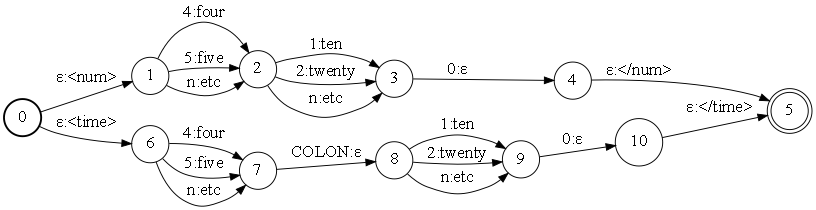}
%   \vspace{-7mm}
  \caption{Tag-Outputting TN FST Network for Training Data}
  \label{fig:tnfst}
%   \vspace{-5mm}
\end{figure}

Training data for the tagger is prepared with the help of Text Normalization (TN) FSTs. These are essentially mirror images of the ITN FSTs discussed in Section \ref{proposed-approach}, but can encode more complicated rules, since memory is not a concern. In addition to converting written form to spoken form, the TN FSTs also output category-tag labels. For \textit{e.g.}, it will convert ``\$25.00 please" to ``\texttt{\texttt{\textless{}money\textgreater{}} twenty five dollars \texttt{\textless{}/money\textgreater{}} please}". A simple example of such a TN FST is in Fig \ref{fig:tnfst}. To generate the training data for the tagger, we form paired data out of the XML strings that are generated by TN FSTs. For \textit{e.g.}, the above XML string will yield, \texttt{[twenty, five, dollars, please]} $\rightarrow$ \texttt{[money, \textunderscore money, \textunderscore money, blank]} as a training sequence. The set of tagging categories we have chosen loosely correspond to the set ITN categories we monitor. 

% Training data for the tagger is prepared with a form of TN based on pattern matching and rewrite rules, implemented via a WFST (see Fig. \ref{fig:tnfst} for a simple example), which outputs category tag labels for mapped strings--essentially, mirror image of the ITN FST used in the runtime.  For example, "\$25.00 please" will become "\textless{}money\textgreater{} twenty five dollars \textless{}/money\textgreater{} please", and converted into vectors of words and features for tagger training, i.e. "twenty,five,dollars,please" + "money,\_money,\_money,-".  

% We have chosen a set of tagging categories aiming for the smallest set needed to distinguish between different mappings of the same input strings, currently: num (natural numbers), ord (ordinals), money, phn (phone numbrers), post (postal codes), time, url, and alnum (nominally "alphanumeric" but really just a generic label of "ITNable").

Many expressions can be mapped to multiple possible lexical forms by competing FST paths, sometimes producing different category tags.  Several mechanisms are used to ensure that TN output takes advantage of these alternatives to reflect the variety of spoken language. An n-gram model, trained on lexical transcriptions, is used to encourage a choice of idiomatically/contextually appropriate translations. For instance, \texttt{101} is mapped to \texttt{\textless{}alnum\textgreater{} one oh one \textless{}/alnum\textgreater{}} in ``hwy 101", but \texttt{\textless{}num\textgreater{} one hundred and one \textless{}/num\textgreater{}} before ``dalmatians".  For some categories, alternate paths can produce different outputs interchangeably; here, the choice is made pseudorandomly: e.g., ``1:45" can become \texttt{\textless{}time\textgreater{} one forty five \textless{}/time\textgreater{}} or \texttt{\textless{}time\textgreater{} quarter to two \textless{}/time\textgreater{}}.  This ensures that the tagger will learn to identify both types as time expressions.

\subsection{Evaluation} \label{sec:evaluation}

We conduct 2 types of evaluations for our models. For the models where ITN is learnt separately, we first perform a ``text-only" evaluation. Our test set for this contains lexical sentences and corresponding display-format. This set allows us to test all models except the display-format ASR model. This test set is rich with ITN instances and was internally developed just to test ITN performance. It contains $\sim$6k utterances with $\sim$11k ITN occurrences overall. We also prepared a ``speech-to-display-format" test set by passing the lexical sentences from the text-only test set through a TTS system and generating speech audio. We use this test set to conduct a more comprehensive and realistic evaluation.

The main metric we track to evaluate ITN performance is Precision/Recall/F1 over the ITN instances in the references. For ``speech-to-display-format" evaluation, in addition to Precision/Recall/F1, we also compute Token Error Rate (TER). TER is simply edit distances computed over space separated display-format output \& display-format reference.

% The list of these ITN categories are present in Table \ref{tab:itn_categories}. 
% Please note, even though we track individual performance over various categories, in this paper we only report aggregate numbers due to limited space. 

% Hence, we show a single Precision/Recall/F1 score for a model candidate in the next sections. Please also note, for ``speech-to-text" evaluation, in addition to Precision/Recall/F1, we also compute Token Error Rate (TER). TER is simply edit distances computed over space separated display-format output \& reference.

% \begin{table}[t]
%   \caption{All instances of ITN are organized under 16 major categories.}
%   \label{tab:itn_categories}
%  \vspace{3mm}
%   \centering
%   {\footnotesize
%   \begin{tabular}{c|l}
%     \hline
%  &  alpha-numeric, date, decimal, email, fraction, math, \\
% ITN categories & money, numbers, ordinals, phone, plural numbers, \\
%  & post codes, roman, time, URL, words\\ \hline

%   \end{tabular}
%   }
% %   \vspace{-3mm}
% \end{table}

\section{RESULTS}\label{sec:results}

% For ITN models that can be learnt independently of ASR, we used the ``text-only" test set described in Section \ref{sec:evaluation}. Based on our results of this round of evaluation, we select a few candidates for a second round of evaluation, which uses the ``speech-to-text" test set to get the full picture.

\subsection{Text-only evaluation} \label{sec:t2t_results}

\begin{table}[t]
  \caption{Evaluation of various ITN models on the lexical-to-display text-only ITN test set. For neural models and components, the size is take after INT8 quantization.}
  \label{tab:t2t_results}
 \vspace{3mm}
  \centering
  {\footnotesize
  \begin{tabular}{lc|ccc}
    \hline
Model &  Size (MB) & Precision & Recall & F1 \\ \hline
WFST + n-gram & 30  & 0.80 & 0.83 & 0.81 \\ 
WFST + n-gram + LSTM & 60  & 0.83 & 0.84 & 0.83 \\ 
S2S-small & 6  & 0.75 & 0.76 & 0.76 \\ 
S2S-large & 82.5  & 0.77 & 0.78 & 0.77 \\ 
Tagger + WFST & 5.5  & 0.81 & 0.84 & 0.82 \\ \hline

  \end{tabular}
  }
  \vspace{-3mm}
\end{table}

Table \ref{tab:t2t_results} shows the results of various ITN models on the `Text-only' evaluation. The first 2 rows correspond to the models described in Section \ref{sec:model_details_wfst} and \ref{sec:wfst}. The ranker in the 1st row is an n-gram and in the second row is a combination of n-gram \& LSTM. It can be seen that `WFST+rescoring' framework is really effective and forms a very strong baseline. Just rescoring the WFST output with an n-gram is able to produce one of the best results in this study. Additional rescoring with an LSTM further improves the results. However, please also note that LSTM rescoring might not be a feasible option in a streaming scenario.

The next 2 results are from seq2seq models trained on parallel lexical \& display-format data. These results correspond to the models described in Section \ref{sec:itn-s2s} \& \ref{sec:model_details_s2s}. With such small footprint, and no human engineered rules, `S2S-small' is able to achieve a respectable F1 of 0.76. It was natural for us to wonder, if simply scaling the model size would help here. `S2S-large' is a proportionally scaled, much larger version. It is clear that simply scaling does not help much. We attribute this to 2 observation: \textit{(i)} real-world ITN is very diverse and even a decently large dataset will likely not have sufficient coverage to cover all ITN patterns. \textit{(ii)} S2S models might be too flexible \& lack robustness for this task. During our error analysis, we did observe a few instances where S2S would ``hallucinate" output \cite{8268937} or rewrite parts of the utterances that shouldn't have been changed. 

The last entry in the table is for our proposed approach. The first thing to notice is that overall model size is 5.5 MB only. 3.5 MB of this comes from the tagger and the remaining 2MB comes from the FST for conversion. A much simpler FST is possible here than with the with the pure-FST approach, or the the FST + rescoring approach, because the FST is only responsible for transformations, not for modeling their contexts. When compared to the WFST + rescoring solutions, we see that it can achieve very similar F1 scores, with a model size that is drastically smaller (in one case less than 10\%). When compared to S2S models, the F1 score of the proposed approach is significantly better, even though they are similar in size.

\begin{table}[t]
  \caption{Relationship between tagger's chunk size, tagger's average algorithmic latency \& model F1 for the proposed approach. Both chunk size and average tagger latency are in number of subword tokens.}
  \label{tab:chunk_size_latency_accuracy}
 \vspace{3mm}
  \centering
  {\footnotesize
  \begin{tabular}{cc|ccc}
    \hline
Chunk size &  Average latency & Precision & Recall & F1 \\ \hline
11 & 5  & 0.81 & 0.84 & 0.82 \\ 
6 & 2.5  & 0.79 & 0.82 & 0.81 \\ 
4 & 1.5  & 0.76 & 0.80 & 0.78 \\ 
2 & 0.5  & 0.72 & 0.77 & 0.74 \\ 
1 & 0  & 0.60 & 0.70 & 0.65 \\ \hline

  \end{tabular}
  }
%   \vspace{-3mm}
\end{table}

Table \ref{tab:chunk_size_latency_accuracy} shows the results for trade-off described in Section \ref{proposed-approach}. We can clearly see that decreasing the chunk size, and consequently the tagger's average latency has a negative effect on the F1 scores. However, the decrease in F1 performance in not significant. For example, going from chunk size of 11 to 6, we can reduce our average tagger latency in half with only a decrease of 1 point in the F1 score. Depending on the application's requirements, the proposed architecture allows us to trade-off latency for ITN performance.

\subsection{Speech-to-text evaluation}

Table \ref{tab:2_stage_result} shows the final ITN performance going from speech to display-format. The first baseline to consider is `Display-format ASR'. This model is trained in an E2E manner to go from speech to display-format transcription directly. As Table \ref{tab:2_stage_result} shows, this model performs the worst of all the candidates evaluated. We mainly attribute this to insufficient ITN variety \& instances in the ASR training data. This is because ITN coverage in not considered in curation of most ASR datasets. The inability of this method to properly leverage our text-only ITN training assets is another contributing factor.

\begin{table}[t]
  \caption{2-stage evaluation with lexical ASR. Size is only shown for the ITN component and does not include ASR.}
  \label{tab:2_stage_result}
 \vspace{3mm}
  \centering
  {\footnotesize
  \begin{tabular}{lc|c|c}
    \hline
Model &  Size (MB) & P/R/F1 & TER \\ \hline
Display-format ASR & -  & 0.63/0.64/0.63 & 25.33 \\ \hline
Lexical ASR &  &  &  \\ 
+ WFST \& n-gram & 30  & 0.72/0.74/0.73 & 22.25 \\ 
+ WFST \& n-gram \& LSTM & 60  & 0.70/0.71/0.71 & 22.00 \\ 
+ Tagger \& WFST \textit{(proposed)} & 5.5  & 0.71/0.75/0.73 & 22.70 \\ \hline

  \end{tabular}
  }
%   \vspace{-3mm}
\end{table}

To test the solutions where ITN is modeled separately, we first decode the speech using a lexical ASR and then apply corresponding ITN solution on it. It is clear, from both F1 scores and TER, that these modular pipelines perform much better than the completely E2E approach. Please see that our tagger approach performs similar to the `WFST + n-gram' solution, in terms of the F1 scores. This is despite being only one-sixth in size. The TER numbers are also similar to each other but please see that we pay more attention to the F1 score. While there is correlation between F1 and TER, this correlation is not perfect. There are cases where TER punishes missed conversion much more than F1. For instance, if we miss converting a phone number, F1 would register missing only 1 ITN instance but a TER is going to register 10 errors.

Next, we consider WFST solution, reranked using both n-gram and LSTM. In Table \ref{tab:t2t_results}, we saw this is supposed to be a superior solution to WFST reranked with n-gram alone. However, as seen in Table
\ref{tab:2_stage_result}, that trend might not hold in an end to end evaluation. Comparing the 2\textsuperscript{nd} and 3\textsuperscript{rd} row in Table \ref{tab:2_stage_result}, additional LSTM rescoring does not improve the F1 score, rather it deteriorates it very slightly. We think this might be because the LSTM was trained on clean text, while output of ASR is not clean and can contain errors. 

\subsection{Runtime Advantages}

\begin{figure}[t]
  \centering
  \includegraphics[width=\linewidth]{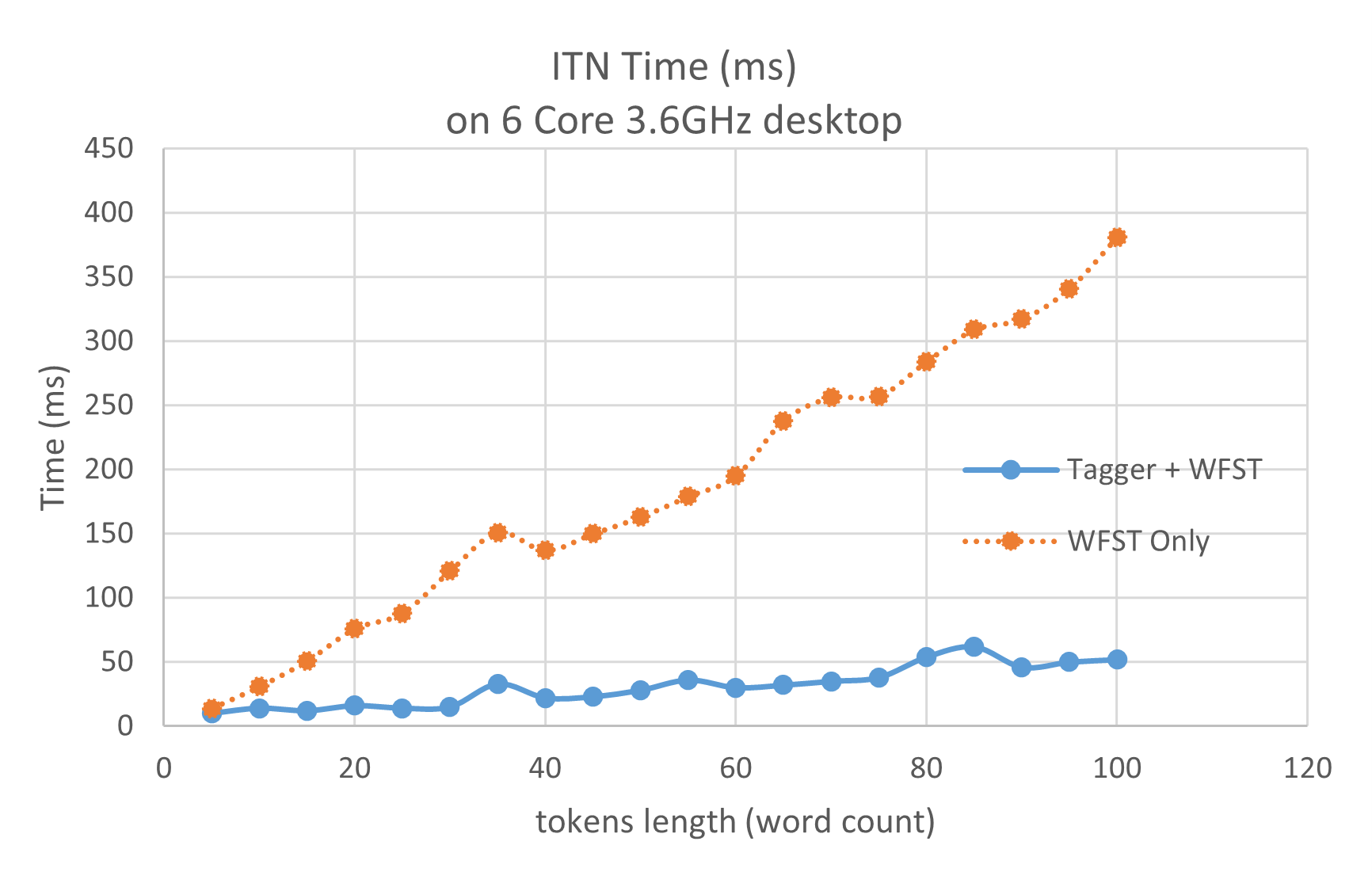}
%   \vspace{-7mm}
  \caption{The proposed approach is runtime friendly.}
  \label{fig:runtime}
%   \vspace{-5mm}
\end{figure}

Consider the WFST+n-gram system. Implementation wise, it makes sense to integrate the n-gram into FST itself. Hence, in this section, we refer to it as the ``WFST-only" solution. Figure \ref{fig:runtime} shows how the runtime computation cost varies for WFST-only system \& our proposed approach, on input sequences of varying length. We see the WFST-only solution's computation cost increases much faster than our tagger-based solution and that for lengthy sequences, it becomes impractical to use. This is because amortized runtime complexities of WFST composition \cite{fst_compose, Mohri2004} do not scale well. While our proposed solution also has a WFST component, that WFST is much smaller and it is only applied on the tagged span, which never exceeds more than 10 tokens. Also note that it easy to accelerate the transformer tagger by batch processing N tokens, while a WFST-only solution has to go with token by token. Besides computation cost, there are several other advantages for the proposed solution, when its deployed with an ASR system. Firstly, we can cache and reuse the results for an ITN span. Since the hypothesis present in the beam search of an E2E ASR are very similar to each other, it is possible that the same ITN span might show up for conversion several times. In such a case, the per-span ITN conversion results have a lot of chance to be reused. Secondly, the transformer tagger solution proposed in this work is naturally streamable. While it's also possible to do stream processing with a WFST, the space and computation cost for it will be a lot higher.

\section{CONCLUSION \& FUTURE WORK}\label{sec:typestyle}

In this paper we propose a novel modeling solution that enables on-device, fast, streaming \& high quality Inverse Text Normalization. Our proposed solution is a ``hybrid" model that leverages the strong contextual prediction of a transformer-tagger and the robustness \& customization abilities of a WFST. We benchmark this approach against several strong baselines and show that our proposed solution can perform ITN on par with best alternative solutions, while being an order of magnitude smaller and having several runtime benefits. In future work, we will extend the tagger's responsibilities to simultaneously learn other post-processing tasks like disfluency removal, punctuation \& capitalization. We will showcase it's effectiveness on both E2E \& Hybrid ASR. We will also explore pre-training schemes and better neural architectures for the tagger, to further improve robustness.
% Below is an example of how to insert images. Delete the ``\vspace'' line,
% uncomment the preceding line ``\centerline...'' and replace ``imageX.ps''
% with a suitable PostScript file name.
% -------------------------------------------------------------------------

% \begin{figure}[htb]
	
% 	\begin{minipage}[b]{1.0\linewidth}
% 		\centering
% 		\centerline{\includegraphics[width=8.5cm]{image1}}
% 		%  \vspace{2.0cm}
% 		\centerline{(a) Result 1}\medskip
% 	\end{minipage}
% 	%
% 	\begin{minipage}[b]{.48\linewidth}
% 		\centering
% 		\centerline{\includegraphics[width=4.0cm]{image3}}
% 		%  \vspace{1.5cm}
% 		\centerline{(b) Results 3}\medskip
% 	\end{minipage}
% 	\hfill
% 	\begin{minipage}[b]{0.48\linewidth}
% 		\centering
% 		\centerline{\includegraphics[width=4.0cm]{image4}}
% 		%  \vspace{1.5cm}
% 		\centerline{(c) Result 4}\medskip
% 	\end{minipage}
% 	%
% 	\caption{Example of placing a figure with experimental results.}
% 	\label{fig:res}
% 	%
% \end{figure}

% \section{REFERENCES}
% \label{sec:ref}
% brackets as shown at the end of this sentence \cite{C2}. An

\bibliographystyle{IEEEbib}
\bibliography{strings,refs}

\end{document}